\pdfoutput=1
\documentclass{article}
\usepackage{spconf,amsmath,graphicx}
\usepackage{subcaption,wrapfig,booktabs,fancyhdr,amsfonts,cite,bm,amssymb,amsthm,url,multirow,times,enumitem,comment}
\newcommand{\ba}{\begin{array}}
\newcommand{\ea}{\end{array}}

\renewcommand{\Re}{\mathbb{R}}

\DeclareMathAlphabet{\mathpzc}{OT1}{pzc}{m}{it}
\title{Automotive Collision Risk Estimation under Cooperative Sensing\vspace{-2ex}}

\name{\normalsize Daniel La{C}hapelle$^{\star}$ \qquad Todd Humphreys$^{\star}$ \qquad Lakshay Narula$^{\dagger}$ \qquad Peter Iannucci$^{\star}$ \qquad Ehsan Moradi-Pari$^{\ddagger}$\vspace{-2ex}}
\address{\small $^{\star}$Aerospace Engineering and Engineering Mechanics, The University of Texas at Austin.\vspace{-1ex}\\ 
         \small $^{\dagger}$Electrical and Computer Engineering, The University of Texas at Austin. \vspace{-1ex}\\ 
         \small $^{\ddagger}$Automobile Technology Research Division, Honda R\&D Americas, Inc.} 

\begin{document}
\ninept
\maketitle
\begin{abstract}


  This paper offers a technique for estimating collision risk for automated
  ground vehicles engaged in cooperative sensing. The technique allows
  quantification of (i) risk reduced due to cooperation, and (ii) the
  increased accuracy of risk assessment due to cooperation. If either is
  significant, cooperation can be viewed as a desirable practice for meeting
  the stringent risk budget of increasingly automated vehicles; if not, then
  cooperation---with its various drawbacks---need not be pursued. Collision
  risk is evaluated over an ego vehicle's trajectory based on a dynamic
  probabilistic occupancy map and a loss function that maps collision-relevant
  state information to a cost metric.  The risk evaluation framework is
  demonstrated using real data captured from two cooperating vehicles
  traversing an urban intersection.
\end{abstract}
\section{Introduction}\label{sec:intro}
Recent works on automated vehicle (AV) safety target a many-fold safety
advantage over human drivers \cite{shalev2017formal,reid2019localization}. To
meet such high expectations, AVs must be designed to continuously and
accurately assess their collision risk over a short time horizon. Cooperative
sensing (inter-vehicle sensor data sharing and drawing sensor data from
infrastructure) may significantly reduce collision risk in highly uncertain,
dynamic environments~\cite{wang2018performance}, but its benefits have yet to
be quantified. Meanwhile, serious challenges to cooperation exist: how can it
be incentivized? Can the ego vehicle trust received data? How can data received 
be accurately placed within the ego vehicle reference frame?

This paper presents a framework for measuring the benefit of cooperation in
terms of reduced collision risk and reduced risk uncertainty. Existing risk
assessment schemes either assume all current and near-future traffic
participants have been identified and are being tracked
\cite{broadhurst2005monte,althoff2009model,lefevre2014survey,carvalho2015automated},
or they assume a worst-case model in which blind spots are fully occupied
\cite{hoermann2017entering}. Neither approach is well suited to an accurate
measurement of risk that accounts for risk lurking in sensing blind spots with
\emph{a priori} probabilities informed by historical sensing. Moreover,
existing approaches do not provide a probability distribution for assessed
risk. This prevents an evaluation of risk uncertainty, and, relevant to the
current paper's aims, an evaluation of the benefit of cooperative sensing in
terms of reduced risk uncertainty.

Predicting the likely evolution of a traffic situation can be performed with an
object-based or an object-free representation of the 
environment~\cite{hoermann2017entering}. An object-based representation assigns a class
(e.g., pedestrian, cyclist, car) to each traffic participant---tracked or
potential---and assumes a class-specific motion model for each
~\cite{broadhurst2005monte,althoff2009model,lefevre2014survey,carvalho2015automated}. 
An object-free representation does not attempt to classify other traffic
participants, but only estimates current and
near-future occupancy in a gridded map. To establish a proof of concept, the current
paper adopts an object-free representation, which requires only a small number
of parameters to model. This can later be extended to an object-based
representation for greater risk assessment accuracy~\cite{Hoermann2018automatic}.

Probabilistic occupancy maps (POMs, also referred to as occupancy grid maps)
are a convenient object-free construct for fusing sensing data from multiple
sensors, possibly from multiple cooperating agents. A POM discretizes the
traffic environment into a grid~\cite{thrun2005probabilistic}. To each of the
grid's cells is associated a Bernoulli random variable: the cell is occupied
with probability $p$ and empty with probability $1-p$. Early POMs represented
only static environments, whereas AV applications require dynamic maps. In
either case, the state of the map at a particular epoch represents the
instantaneous estimates of occupancy probabilities conditioned on all previous
measurements. Methods for estimating occupancy over time from noisy
measurements are known in the literature as Bayesian occupancy filters 
(BOFs)~\cite{saval2017review}. At each point in time, the state of the BOF is a
POM. Since it was introduced for automotive motion 
planning~\cite{coue2006bayesian}, BOF theory has been developed in a variety of
different ways~\cite{saval2017review}. Some approaches are optimized for
particular sensing systems~\cite{perrollaz2012visibility}, whereas others
reduce errors by incorporating prior map knowledge~\cite{gindele2009bayesian}.

The key advantage of BOFs over traditional multi-target tracking is that BOFs
sidestep the notoriously thorny combinatorial ``data association problem'' in
which sensor measurements must be associated with a discrete observed target~\cite{fan2017dual}. The BOF and related approaches do not associate sensor
measurements explicitly. Instead, measurements are used to update the
occupancy probability of grid cells under the assumption that the cells are
statistically independent. For BOF proponents, discretization and assumed 
cell-to-cell statistical independence, although introducing some inaccuracy, 
are justified by convenience:  they allow Bayesian update equations to be 
represented analytically and updated in parallel~\cite{danescu2011modeling, 
negre2014hybrid, rummelhard2015conditional, steyer2019grid}.

More recent dynamic mapping approaches employ particle filter tracking to ease
the computational burden of BOFs. In
particular, the Probability Hypothesis Density / Multi-Instance Bernoulli
(PHD/MIB) filter~\cite{nuss2018random} is at the core of the current
paper's risk assessment formulation. It is particularly attractive because 
it links the problem of dynamic state estimation of grid cells to a 
rigorously-founded literature in finite set statistics 
(FISST)~\cite{vo2013labeled}, ensuring that particles have a well-defined
physical meaning and are propagated according to Bayesian rules.
A BOF approach rooted in random finite set theory enables risk estimation 
to incorporate not only the risk of collision with observed objects, but also 
with objects that have not been observed but could possibly exist, informed by historical data.

This paper makes two primary contributions. First, it develops a technique for
collision risk evaluation that is amenable to data fusion from multiple
cooperating vehicles (or from infrastructure), and amenable to realistic
characterization of risk emerging from occluded areas (i.e., blind spots).
Besides risk evaluation, the technique also offers a measure of risk
uncertainty, which one would expect to decrease under cooperative sensing.
Second, the paper applies the risk assessment framework to real data captured
from a pair of connected vehicles to demonstrate the benefit of
cooperative sensing for a simple example scenario.

\section{Approach}\label{sec:approach}
The core contribution of this paper is a framework for computing estimated
collision risk based on a Bayesian occupancy filter augmented with a loss
function. The BOF, along with the ego vehicle's pose estimate, can be used to
compute the probability of collision at time step $k$ in grid cell $c$; the
loss function returns the severity of the collision. This paper presents
accumulated risk, a quantity evaluated over a given trajectory: the
accumulated risk at time step $k$ is the cumulative sum of risks up to $k$.

\subsection{Bayesian Occupancy Filtering}\label{sec:map}
To implement a Bayesian occupancy filter, this paper adopts the PHD/MIB formalism introduced
by Nuss et al. \cite{nuss2018random}. In this approach, each grid cell is
modeled as an independent Bernoulli random finite set (RFS) that is either
empty with probabilty $1-r$ or contains a single point object with 
existence probability $r$ and spatial distribution $p(\mathbf{x})$. In 
the two-dimensional problem, $p(\mathbf{x})$, $\mathbf{x}\in\Re^4$ 
is the density of the point object's two-dimensional position and velocity.
In other words, if $X$ is the RFS for cell $c$, then $X$ can be either 
$\mathbf{x}$ or $\emptyset$, and its FISST density is given by
\begin{equation}\label{eq:rfs}
\pi_c(X) = \begin{cases} 1-r                  & \text{if } X=\emptyset \\ 
                         r\cdot p(\mathbf{x}) & \text{if } X=\{\mathbf{x}\}
              \end{cases}
\end{equation}
In this application, $r$ is the occupancy probability of cell $c$ with FISST density $\pi_c(X)$.  The notion
of a grid cell containing a single point object may seem like a crude
approximation, especially if grid cells are large enough to contain multiple
different real-world objects. However, it is implicitly the same
approximation that other occupancy grid approaches make. An object in the real
world will, when represented in the map, consist of one or more of these point
objects depending on its spatial extent and the grid discretization.

The PHD/MIB filter for the dynamic occupancy map is realized by a
set of weighted particles to model $p(\mathbf{x})$. The occupancy probability
of a given cell can be easily recovered: it is simply the sum of weights of
the particles associated with that cell's point object.

\subsection{Measurement Models}\label{sec:meas}
Just as there are two possible truth states for each grid cell (empty and
occupied), so too are there two possible information-yielding sensor measurements for each grid
cell: ``detection'' and ``miss.'' Sensors for probabilistic occupancy mapping
may be modeled as beams that extend from the sensor until they hit a solid
object in their field of view. This model closely mirrors reality for common
sensors like radar, lidar, and stereo cameras. A detection occurs for the cell
in which the sensor returns a range less than its maximum range; a miss occurs
for all cells in between that cell and the sensor, along the radial direction.

This binary measurement model can be challenging to wield in the presence of
sensor pose and measurement uncertainty---how can one be certain that a detection falls in a
given cell versus the one adjacent? For this reason, this paper implements a
Monte Carlo beam measurement model in which each measurement beam is
represented by a set of beams randomly sampled from a distribution that takes
pose and measurement uncertainty into account. Thus, the measurements at epoch
$k$ take the form of two arrays: one containing the expected number of
detections in each cell, and the other containing the expected number of
misses.


Nuss et al. model measurements in each grid cell as an independent Bernoulli
observation process, in which each grid cell either has one observation or no
observation. However, when pose and sensor uncertainty is non-negligible, as
it is in many applications, it is likely that a grid cell will have multiple
conflicting observations. In this sense, repeated Bernoulli experiments for a
given grid cell at a given point in time are better modeled as a Binomial
random variable. Because the chief interest is in estimating $r$, the
occupancy probability of the cell, the natural choice is to model it using a
Beta distribution.
\begin{equation}\label{eq:beta}
r\sim f(\rho;\alpha, \beta) = \frac{\rho^{\alpha-1}(1-\rho)^{\beta-1}}{B(\alpha, \beta)}
\end{equation}
\begin{equation}\label{eq:exp}
\mathbb{E}[r] = \frac{\alpha}{\alpha+\beta}
\qquad
Var[r] = \frac{\alpha\beta}{(\alpha+\beta)^2(\alpha+\beta+1)}
\end{equation}
This model has numerous benefits. It is fully parameterized by two numbers, $\alpha$ and $\beta$, and it can be used to compute the variance of the occupancy probability estimate, yielding more information without a much greater computational burden. In general, Bayes theorem for this problem can be written as
\begin{equation}\label{eq:bayes}
g(r|Z_1, \ldots, Z_n) = \frac{g(r) \prod_{i=1}^n h(z_i|r)}{\int_\Omega g(r) \prod_{i=1}^n h(z_i|r) dr} 
\end{equation}
In this expression, $g(r)$ is the Bayesian prior and $\prod_{i=1}^n h(z_i|r)$ is the likelihood; it yields the posterior distribution $g(r|Z_1, \ldots, Z_n)$. The integral in the denominator of Equation \ref{eq:bayes} must be computed numerically. However, as the Beta distribution is a member of the exponential family of distributions, a Beta prior yields a Beta posterior if Bernoulli sampling is used \cite[Ch. 13]{gupta2004handbook}. More specifically, for a prior distributed as $\mathrm{Beta}(\alpha, \beta)$, the posterior after Bernoulli sampling with $k$ ``successes'' and $m$ ``failures'' is distributed as $\mathrm{Beta}(\alpha+k, \beta+m)$. A reasonable choice for a prior to start with for each cell is the $\mathrm{Beta}(1,1)$ distribution: the uniform distribution on $[0, 1]$. Note that the expected value of $\mathrm{Beta}(1,1)$ is $\frac{1}{2}$, the typical starting point of a cell's occupancy probability in a Bayesian occupancy filter in the absence of prior information.

It should also be noted that the Bernoulli random variable modeling the occupancy of the cell is not the same as the Bernoulli random variable modeling the detections and misses from that cell. Different sensors have different properties; a detection in a cell from one sensor may be more likely than one from another. It is reasonable, however, to assume that the two are related. This paper suggests defining pseudo-detections and pseudo-misses as functions of the expected sensor detections and misses. Given a Beta-distributed prior for occupancy probability,$f(\rho; \alpha, \beta)$, one can assume that the posterior takes the form $f(\rho; \alpha + \Delta\alpha, \beta + \Delta\beta)$. One can think of $\Delta\alpha$ as the number of pseudo-detections and $\Delta\beta$ the number of pseudo-misses. A reasonable assumption is that $\Delta\alpha$ and $\Delta\beta$ are functions of the detections and actual misses, respectively. Moreover, one is interested in fusing the detections and misses from heterogeneous sensors. This paper models the pseudo-detections as linear combinations of actual expected detections; that is,
\begin{equation}\label{eq:weights}
\Delta\alpha = \sum_{s=1}^S w_s^{\mathrm{det}}n_s^{\mathrm{det}}
\qquad
\Delta\beta = \sum_{s=1}^S w_s^{\mathrm{miss}}n_s^{\mathrm{miss}}
\end{equation}
over an index set of sensors $S$, where $w_s^{\mathrm{det}}$ ($w_s^{\mathrm{miss}}$) is the weight associated with detections (misses) from sensor $s$ and $n_s^{\mathrm{det}}$ ($n_s^{\mathrm{miss}}$) is the expected number of detections (misses) from that sensor. The weights $w_s^{\mathrm{det}}$ and $w_s^{\mathrm{miss}}$ may be determined heuristically; one suggestion is to compute them using the update equations for a single measurement in a binary Bayes filter framework. Given the occupancy likelihood functions for cell $c$ and sensor $s$ $g_s^{(c)}(z|o)$ and $g_s^{(c)}(z|\overline{o})$ from Nuss et al. \cite{nuss2018random}, where $o$ denotes the event a cell is occupied and $\overline{o}$ the event that it is empty, it is possible to derive the following expressions for the weights for small $\alpha$, $\beta$:
\begin{equation}\label{eq:detweight}
w_s^{\mathrm{det}} \approx \left( \frac{g_s^{(c)}(z=\mathrm{det}|o)}{g_s^{(c)}(z=\mathrm{det}|\overline{o})} - 1\right)\alpha
\end{equation}
\begin{equation}\label{eq:missweight}
w_s^{\mathrm{miss}} \approx \left( \frac{g_s^{(c)}(z=\mathrm{miss}|\overline{o})}{g_s^{(c)}(z=\mathrm{miss}|o)} - 1\right)\beta
\end{equation}
The ratios in these expressions are recognizable as likelihood ratios.

\subsection{Risk Evaluation}\label{sec:risk}
Risk is a measure of expected loss. The expectation, in this case, is taken over the probability that the ego vehicle will collide with with the point object in cell $c$ over the time interval $\tau_k$. Maintaining the assumption that cells are statistically independent, the risk $R_k$ to which the ego vehicle is exposed at time $k$ can be written as
\begin{equation}\label{eq:risk}
R_k = \sum_c r(c, k)\cdot p_{\mathrm{ego}}(c, k)\cdot L(\mathbf{x}_{\mathrm{ego}}(k), \mathbf{x}_c(k))
\end{equation}
where $c$ is the cell index, $r(c, k)$ is the probability that cell $c$ is occupied at epoch $k$, $p_{\mathrm{ego}}(c, k)$ is the probability that the ego vehicle is in cell $c$ at time $k$, and $L(\mathbf{x}_{\mathrm{ego}}, \mathbf{x}_c)$ is the loss as a function of the ego vehicle state $\mathbf{x}_{\mathrm{ego}}$ and the cell state $\mathbf{x}_c$. Both $\mathbf{x}_{\mathrm{ego}}$ and $\mathbf{x}_c$ are random vectors in $\Re^4$ with two-dimensional position and velocity distributions.

\subsection{Loss Function}\label{sec:loss}
Understandably, the resulting risk estimate is driven by the choice of loss function.
This paper proposes a simple loss function based on kinetic energy, where the loss due to collision in a given cell $c$ is written as
\begin{equation}\label{eq:loss}
L(\mathbf{x}_{\mathrm{ego}}(k), \mathbf{x}_c(k)) = C_1\Vert\mathbf{v}_{\mathrm{ego}} - \overline{\mathbf{v}}_c \Vert^2 + C_2 \mathbb{E}[\Vert\mathbf{v}_c - \overline{\mathbf{v}}_c\Vert^2]
\end{equation}
where $\mathbf{v}_{\mathrm{ego}}$ is the expected velocity of the ego vehicle and $\overline{\mathbf{v}}_c$ is the weighted mean of particle velocities in cell $c$. In the second term, $\mathbf{v}_c$ is the random variable of cell $c$'s velocity, part of $\mathbf{x}_c$. The constants $C_1$ and $C_2$ are yet another design choice; in the ``kinetic energy'' case, $C_1 = m_{\mathrm{ego}}m_c/(2(m_{\mathrm{ego}}+m_c))$ and $C_2=m_c/2$ where $m_{\mathrm{ego}}$ is the mass of the ego vehicle in the relevant cell and $m_c$ is the mass of the cell. The cell loss must be normalized by multiplying by the cell area $A_c$ and the time discretization interval $\tau$, ensuring that the risk metric is not sensitive to the choice of discretization.


\section{Experimental Results}\label{sec:results}
\subsection{Data Capture}

\begin{figure}[htb]
\begin{minipage}[b]{1.0\linewidth}
   \centering
   \centerline{\includegraphics[width=8.5cm]{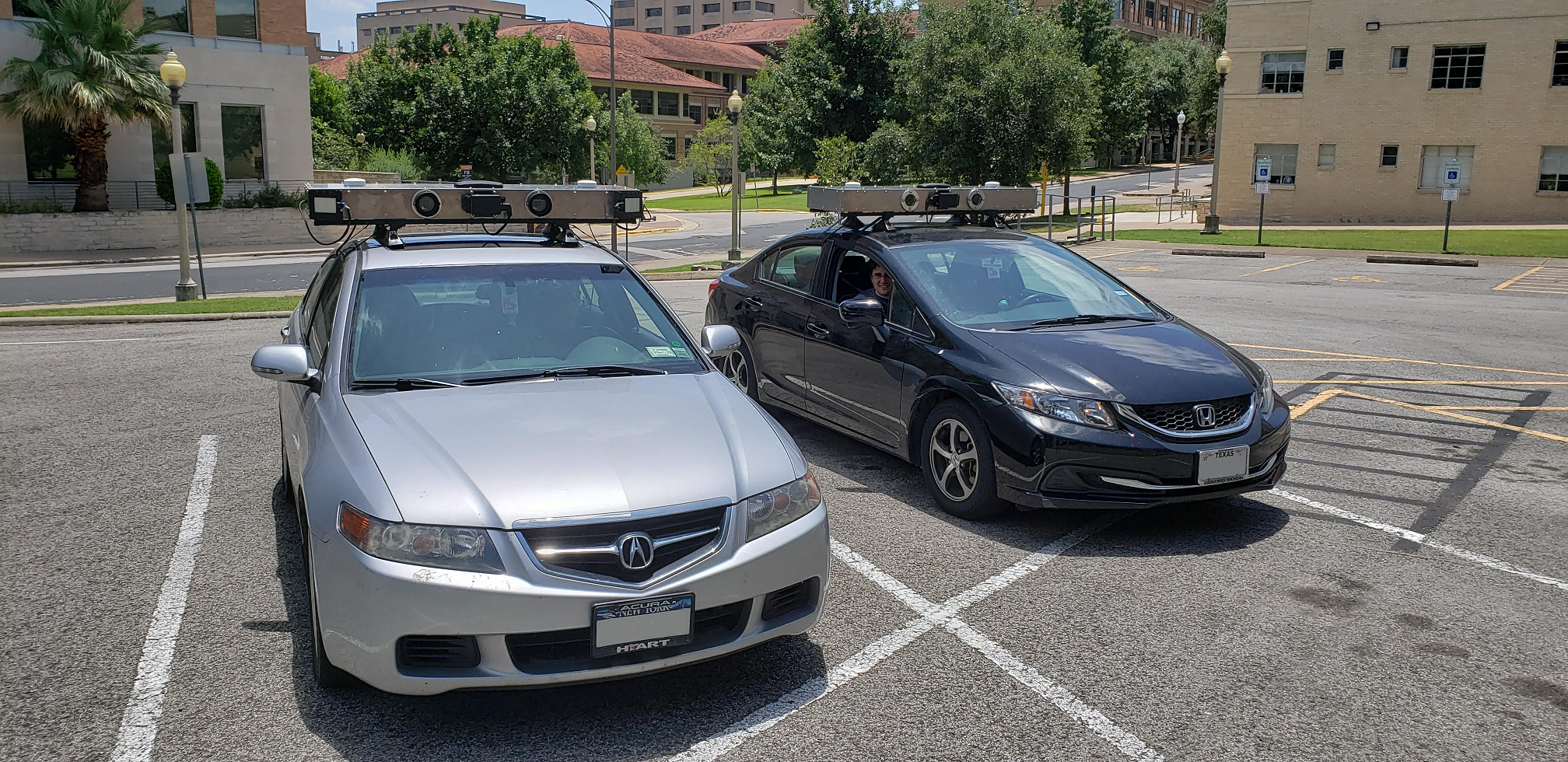}}
\end{minipage}
\caption{The University of Texas \textit{Sensorium} is a platform for automated and connected vehicle perception research. It includes stereo visible light cameras, an industrial grade IMU, one to three automotive radar units, a dual-antenna, dual-frequency software-defined GNSS receiver, 4G cellular connectivity, and a powerful internal computer.}
\label{fig:sensorium}
\end{figure}

\begin{figure}[htb]
\begin{minipage}[b]{1.0\linewidth}
   \centering
   \centerline{\includegraphics[width=8.5cm]{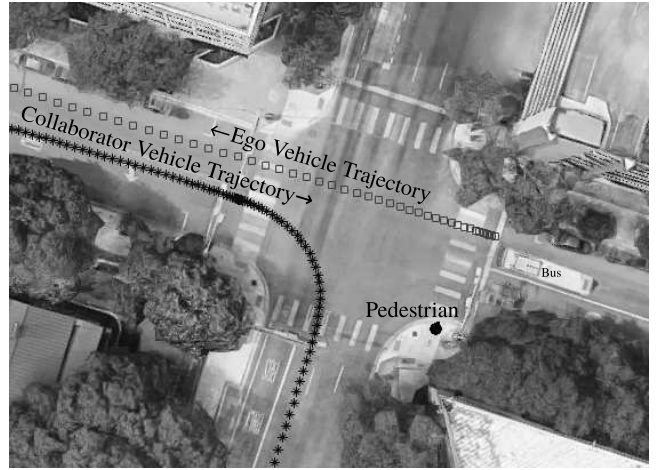}} 
\end{minipage}
\caption{Illustration of the 8th and Guadalupe Scenario.}
\label{fig:cartoon}
\end{figure}

In order to demonstrate the benefits of collaborative sensing, the two University of Texas \textit{Sensorium} sensor platforms (Fig. \ref{fig:sensorium}) were deployed at an intersection in downtown Austin, Texas (Fig. \ref{fig:cartoon}). The data capture was timed such that a bus obscured much of the intersection from the point of view of the vehicle traveling east to west, the ego vehicle. Meanwhile, a collaborator vehicle, traveling in the other direction but turning right (south), had a largely unobstructed view of the entire intersection. Two separate risk evaluations were performed for the ego vehicle: one in which the ego vehicle only has access to its own sensing, and another in which the ego vehicle may also take advantage of sensor information from the collaborator vehicle. 

Centimeter-accurate vehicle poses were computed based on a dual-antenna RTK receiver with MEMS-grade inertial aiding fused in a loosely-coupled sense in an unscented Kalman filter. The sensor measurement used in this work is depth computed from dense stereo disparity maps and projected into a shared local East, North, Up (ENU) coordinate frame.

\begin{figure}[t]
\begin{minipage}[b]{1.0\linewidth}
   \centering
   \centerline{\includegraphics[width=7.5cm]{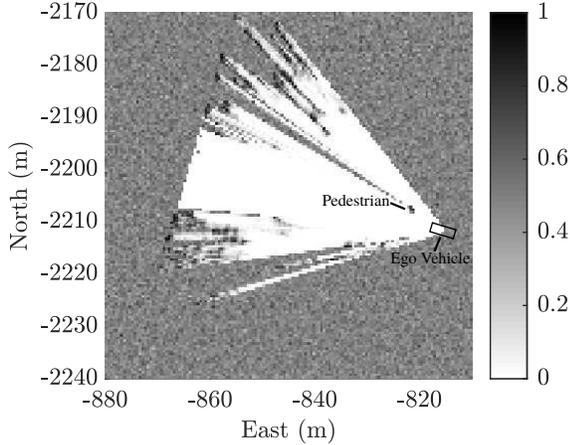}} 
\end{minipage}
\caption{Occupancy map at $t=61.6$ seconds with data from only the ego vehicle.}
\label{fig:ego}
\end{figure}
\begin{figure}[t]
\begin{minipage}[b]{1.0\linewidth}
   \centering
   \centerline{\includegraphics[width=7.5cm]{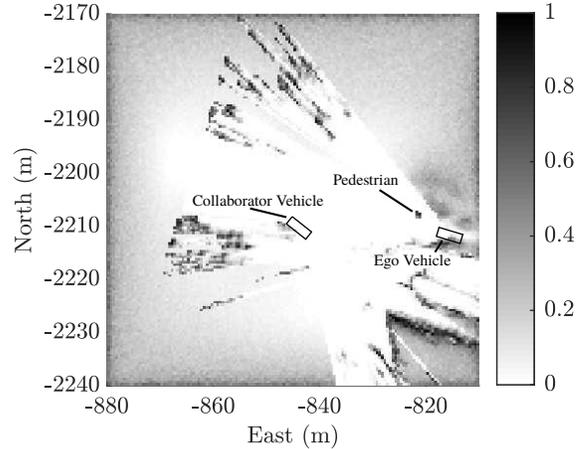}} 
\end{minipage}
\caption{Occupancy map at $t=61.6$ seconds built with data from both vehicles.}
\label{fig:collab}
\end{figure}

Figs. \ref{fig:ego} and \ref{fig:collab} display the dynamic probabilistic occupancy maps at one time instant using ego-only data and combined ego-collaborator data, respectively. Dark-colored cells are more likely to be occupied, while a light-colored cells are more likely to be empty. Fig.~\ref{fig:risk} compares the collision risk of the ego vehicle evaluated using these dynamic probabilistic occupancy maps.

\subsection{Discussion}
Collaborative sensing allows the ego vehicle to effectively see beyond the field of view of its own sensors. In the presented scenario, collaborative sensing enables the ego vehicle to see the pedestrian cross the street while it is obscured by the bus. In addition, collaborative sensing allows the vehicle to more confidently rule out the possibility that there are other objects out of its field of view that pose a risk for collision.

The initial increase in the accumulated risk is due to the ego vehicle sharing its immediate environment with other nearby objects, and its increase in velocity as it accelerates into the intersection. Once in the intersection, however, the ego vehicle is able to combine its sensing with the collaborator vehicle's sensing to verify that the intersection is indeed clear, and the accumulated risk curve flattens. As the vehicle exits the map, the risk increases again as the boundaries of the map are restricted to the ``ignorant'' $\mathrm{Beta}(1,1)$ prior.

\begin{figure}[htb]
\begin{minipage}[b]{1.0\linewidth}
   \centering
   \centerline{\includegraphics[width=8.5cm]{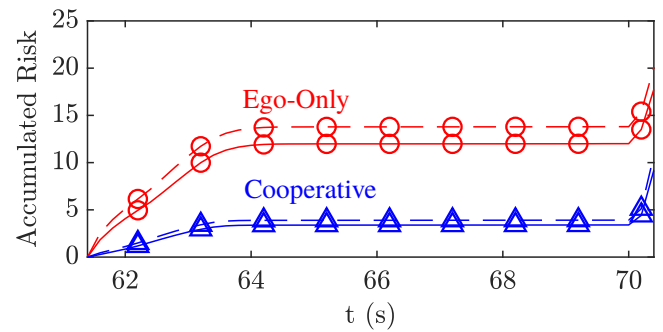}} 
\end{minipage}
\caption{Accumulated risk profiles for the ego vehicle. The solid lines represent expected accumulated risk, while the dashed lines represent the expected accumulated risk plus two standard deviations.}
\label{fig:risk}
\end{figure}

It is clear that the benefit of sensor sharing is reduced collision risk from Fig.~\ref{fig:risk}. It is also worth noting that the collaborative risk estimate is also more confident, as can be inferred from the $+2\sigma$ line. In this case, of course, there was no collision; this estimate of collision risk also takes into account exposure to potential collisions due to objects in unobserved regions of the environment. In a situation where collaborative sensing actually increases the expected collision risk (for example, by warning the ego vehicle about an impending unpredicted collision), the variance of the collision risk estimate should decrease thanks to the increased certainty provided by collaborative sensing.

\section{Conclusions}\label{sec:conc}
This paper presents a framework for automotive collision risk estimation based on a Bayesian occupancy filter augmented with a collision loss function. To test the framework, a data capture was conducted at an intersection in downtown Austin, Texas. The results of the experiment support the notion that cooperative sensing can reduce exposure to collision risk.


This risk framework can estimate collision risk for a known trajectory, as presented in this paper; alternatively, it can be used for motion planning by computing the least risky trajectory given current sensing. The use of cooperative sensing raises additional questions: if communication bandwidth is limited, which sensor data is the most useful? In this way, the motion planning problem can be reframed in terms of active sensing and data exchange. 

\section{Acknowledgments}
This work has been supported by Honda R\&D Americas, Inc. as an affiliate of
The University of Texas Situation-Aware Vehicular Engineering Systems (SAVES)
Center (\url{http://utsaves.org/}), an initiative of the Wireless Networking and
Communications Group. 

\bibliographystyle{IEEEbib.bst}
\bibliography{paper.bib}

\end{document}